\documentclass[conference]{IEEEtran}
\IEEEoverridecommandlockouts
\usepackage{cite}
\usepackage{graphicx}
\usepackage{subcaption}
\usepackage{multirow}
\usepackage{comment}
\usepackage{amsmath,amssymb,amsfonts}
\usepackage{graphicx}
\usepackage{textcomp}
\usepackage{xcolor}
\usepackage{algorithm}
\usepackage{algpseudocode} 
\usepackage[utf8]{inputenc}
\usepackage{varwidth}
\usepackage{bm} 
\usepackage{balance}
\usepackage{soul,color}
 \usepackage{array}
 \usepackage{float}
 \usepackage{svg}
\usepackage{url}
\def\BibTeX{{\rm B\kern-.05em{\sc i\kern-.025em b}\kern-.08em
    T\kern-.1667em\lower.7ex\hbox{E}\kern-.125emX}}
    
\begin{document}

\title{Spatial Transformer Network YOLO Model for Agricultural Object Detection\\

 \author{\IEEEauthorblockN{Yash Vivek Zambre$^1$, Ekdev Rajkitkul$^2$, Akshatha Mohan$^1$, Joshua Peeples$^1$}
 \IEEEauthorblockA{$^1$Department of Electrical and Computer Engineering, Texas A\&M University, College Station, TX, USA\\
	\IEEEauthorblockA{$^2$Department of Computer Science and Engineering, Texas A\&M University, College Station, TX, USA\\
		yashzambre@tamu.edu, ekdev@tamu.edu, akshatha.mohan@tamu.edu, jpeeples@tamu.edu}}}

%\thanks{} (add acknowledgement later)
}
\maketitle

\begin{abstract}
Object detection plays a crucial role in the field of computer vision by autonomously locating and identifying objects of interest. The You Only Look Once (YOLO) model is an effective single-shot detector. However, YOLO faces challenges in cluttered or partially occluded scenes and can struggle with small, low-contrast objects. We propose a new method that integrates spatial transformer networks (STNs) into YOLO to improve performance. The proposed STN-YOLO aims to enhance the model's effectiveness by focusing on important areas of the image and improving the spatial invariance of the model before the detection process. Our proposed method improved object detection performance both qualitatively and quantitatively. We explore the impact of different localization networks within the STN module as well as the robustness of the model across different spatial transformations. We apply the STN-YOLO on benchmark datasets for Agricultural object detection as well as a new dataset from a state-of-the-art plant phenotyping greenhouse facility. Our code and dataset are publicly available \footnote{\url{https://github.com/Advanced-Vision-and-Learning-Lab/STN-YOLO}}.
\end{abstract}


\section{Introduction}
Plant phenotyping is critical for crop improvement \cite{plant_phenotype_crop}, yield optimization \cite{Yield}, and sustainable practices \cite{phenotype}. Artificial intelligence (AI), particularly object detection algorithms, has transformed plant phenotyping, enhancing efficiency and performance \cite{AIinPhenotype}. The You Only Look Once (YOLO) has been used effectively in various Agricultural applications such as pest detection \cite{pest}, crop disease detection \cite{crop}, and crop harvesting \cite{crop_harvest}. Despite the vast number of use cases, YOLO has some limitations due to various spatial transformations \cite{yolo_limitation}. Spatial transformer networks (STNs) \cite{STN} are an approach to improve an artificial neural network's robustness to spatial transformations. We propose to integrate STNs with the YOLO model to incorporate spatial invariance. The STN applies learnable affine transformations to the images that will help with object detection. The STN-YOLO model demonstrates spatial invariance and outperforms the baseline YOLO model on several Agricultural benchmark datasets.

\begin{figure}[t]
	\centering
\includegraphics[width=1\linewidth]{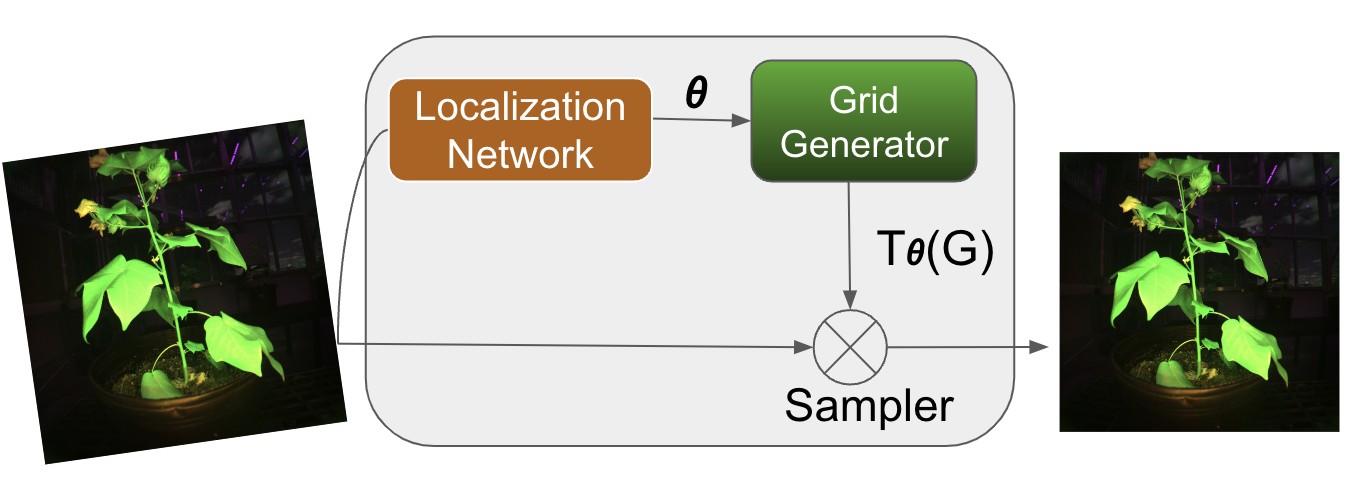}
	\caption[Spatial Transformer Network]{Our object detection framework leverages STNs to improve spatial invariance in the network for object detection. STNs are comprised of three components: localization network, grid generator, and sampler. The localization network outputs affine transform parameters ($\theta$) that are passed to the grid generator. The grid transformation, $T_{\theta}(G)$, is applied to the input image resulting in the new output image.}
	\label{stn}
\end{figure}

Additionally, we present a new, high quality, annotated dataset of various plants to advance the study of agriculture plant detection and phenotyping. The new Plant Growth and Phenotyping (PGP) dataset differs from other available image datasets in the following aspects: (1) multi-spectral images captured across varying heights; (2) challenging images of multiple crops across varying illumination conditions; (3) precise annotations assisted with the use of Segment Anything model \cite{SAM}, (4) large size and shape variations of plants. The features of the PGP dataset present new challenges for agricultural object detection. We perform extensive experiments on the proposed database and new model (STN-YOLO) for object detection. The key contributions of our work are the following:
\begin{itemize}
    \item Integration of STN within YOLO for improved object detection performance
    \item New benchmark dataset for plant object detection.
\end{itemize}
\section{Related Work}
\subsection{Object Detection Models}
Object detection is a prominent challenge in computer vision, where researchers use deep learning to boost performance \cite{parab2022comparison, advances_in_OD1, advances_in_OD2}. Two-stage detectors, such as Faster R-CNN \cite{FASTER_RCNN}, involve distinct stages with a region proposal network (RPN) and region of interest (ROI) pooling for candidate bounding boxes \cite{parab2022comparison}. In contrast, single-stage detectors such as single-shot detectors (SSD) predict bounding boxes directly by utilizing grid boxes and anchors, as described in \cite{SSD}. Notably, YOLO has emerged as a prominent example of a single-stage detector. The advancements of CNNs architectures introduces innovations such as anchor boxes to enhance object detection \cite{advances_in_OD2}. Key advancements include R-CNN, integrating region proposals with CNN using support vector machine (SVM) classification and bounding-box regression \cite{rcnn}. However, R-CNN incurs substantial computational costs and information loss. Fast R-CNN extends this with ROI pooling and proposal refinement, improving test speed and algorithm precision \cite{FAST_RCNN}. Faster R-CNN further reduces computational requirements by introducing the RPN for regional proposals, selecting anchors based on specific criteria for outstanding recognition precision \cite{FASTER_RCNN}. Ongoing advancements focus on speed optimization, including anchor-free detection \cite{an_yolov6}.

\subsection{Spatial Transformer Network (STN)}
\label{sect:STN}
STN is a module with differentiable properties that performs learnable spatial transformation on input feature maps \cite{STN}. The transformation is dependent on the specific input, generating a singular output feature map. In the case of multi-channel inputs, an identical transformation is applied on each channel. The STN module is comprised of three components, as illustrated in the Figure \ref{stn}. The first component is a localization network, which passes input feature maps through several hidden layers to generate the parameters for a learnable affine transformation that is applied to each feature map. Subsequently, the predicted transformation parameters \(\theta\) are used to construct a sampling grid—a set of points indicating where the input feature maps should be sampled to generate the transformed output. Finally, the feature maps and the sampling grid serve as inputs to the sampler that generates the output map sampled from the input at the specified grid points.

\noindent\textbf{Localization Networks} The localization network ($f_{loc}$) takes the input feature maps $\mathbf{X} \in \mathbb{R}^{H \times W \times C}$ with width \(W\), height \(H\), and \(C\) channels. The features extracted by the localization network are then passed into a fully connected layer that outputs \(\theta\), the parameters of the transformation \(T_{\theta}\) to be applied to each feature map: $\theta = f_{loc}(X)$. The size of \(\theta\) can vary depending on the transformation type that is parameterized \cite{STN}. The localization network function \(f_{\text{loc}}()\) can take any form, such as a fully-connected network or a CNN such as ResNet18 \cite{RESNET18}, but should include a final regression layer (\textit{i.e.}, fully connected layer) to produce the transformation parameters \(\theta\) (\textit{e.g.}, affine transformation have six parameters).

\noindent\textbf{Parameterized Sampling Grid}
To perform a transformation on the input feature map, each resulting ``pixel" is determined by applying a sampling kernel centered at a specific location in the input feature map. The term ``pixel" in this context refers to an element of a general feature map, not necessarily an image. In a general context, the resulting ``pixels" are positioned on a regular grid \(G = \{G_i\}\) of pixels \(G_i = (x^t_{i}, y^t_{i})\), creating an output feature map \(V \in \mathbb{R}^{H' \times W' \times C}\). Here, \(H'\) and \(W'\) represent the height and width of the grid, respectively, and \(C\) is the number of channels, which is same in the input and output feature maps.

For clarity of exposition, assume that \(T_{\theta}\) is a 2D affine transformation \(A_{\theta}\). In the affine case, the point-wise transformation is shown in Equation \ref{eq:transformation_matrix}:
\begin{equation}
\begin{bmatrix}
    x_{i}^{s} \\
    y_{i}^{s}
\end{bmatrix}
= T_{\theta}(G_i) = A_{\theta}
\begin{bmatrix}
    x_{i}^{t} \\
    y_{i}^{t} \\
    1
\end{bmatrix}
= 
\begin{bmatrix}
    \theta_{11} & \theta_{12} & \theta_{13} \\
    \theta_{21} & \theta_{22} & \theta_{23} 
\end{bmatrix}
\begin{bmatrix}
    x_{i}^{t} \\
    y_{i}^{t} \\
    1
\end{bmatrix}
\label{eq:transformation_matrix}
\end{equation}

\noindent In this scenario, \((x_{i}^{t}, y_{i}^{t})\) denotes the target coordinates of the regular grid in the output feature map, and \((x_{i}^{s}, y_{i}^{s})\) represents the source coordinates in the input feature map that define the sample points. The relationship between these coordinates is determined by the affine transformation matrix \(A_{\theta}\) \cite{STN}. The STN is self-contained module that is versatile and can be seamlessly integrated at any point in an artificial neural network architecture. The inclusion of an STN at the beginning of a CNN enables the network to learn how to dynamically transform input images to account for spatial variations.

\section{Method}

\subsection{Plant Growth and Phenotyping Dataset}

\begin{figure}[htb]
    \centering
    \setlength{\fboxsep}{0pt} % Remove padding around subfigures
    \begin{subfigure}[b]{0.3\linewidth} % Adjust width of subfigures
        \centering
        \includegraphics[width=1\linewidth]{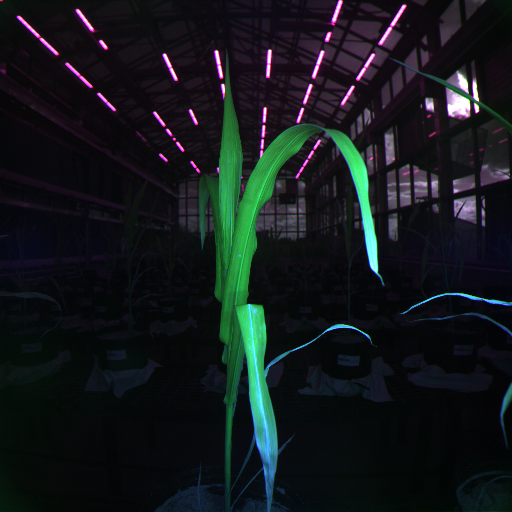}
        \caption{Corn}
        \label{fig:corn_img}
    \end{subfigure}\hfill % Adjust spacing between subfigures
    \begin{subfigure}[b]{0.3\linewidth}
        \centering
        \includegraphics[width=\linewidth]{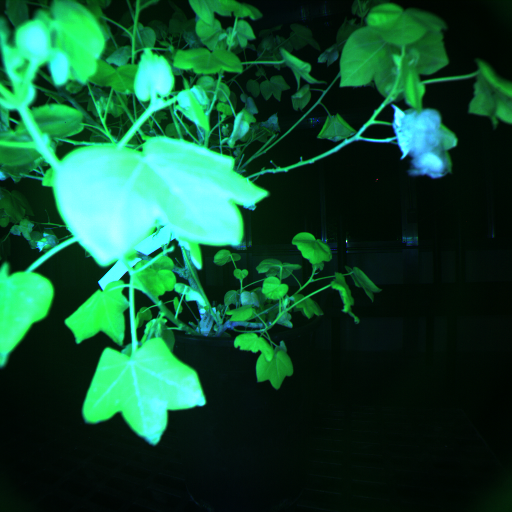}
        \caption{Cotton}
        \label{fig:cotton_img}
    \end{subfigure}\hfill
    \begin{subfigure}[b]{0.3\linewidth}
        \centering
        \includegraphics[width=\linewidth]{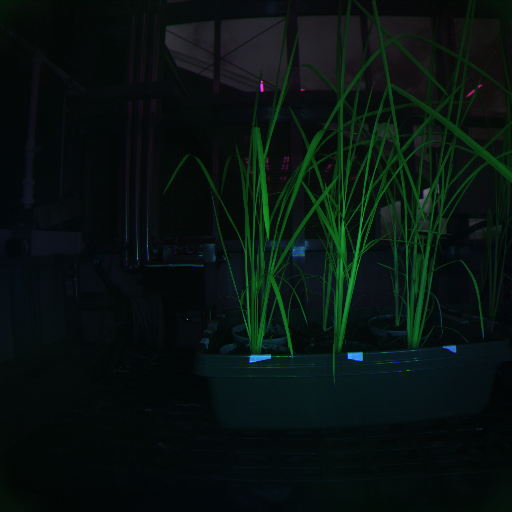}
        \caption{Rice}
        \label{fig:rice_img}
    \end{subfigure}

    \begin{subfigure}[b]{0.3\linewidth} % Adjust width of subfigures
        \centering
        \includegraphics[width=1\linewidth]{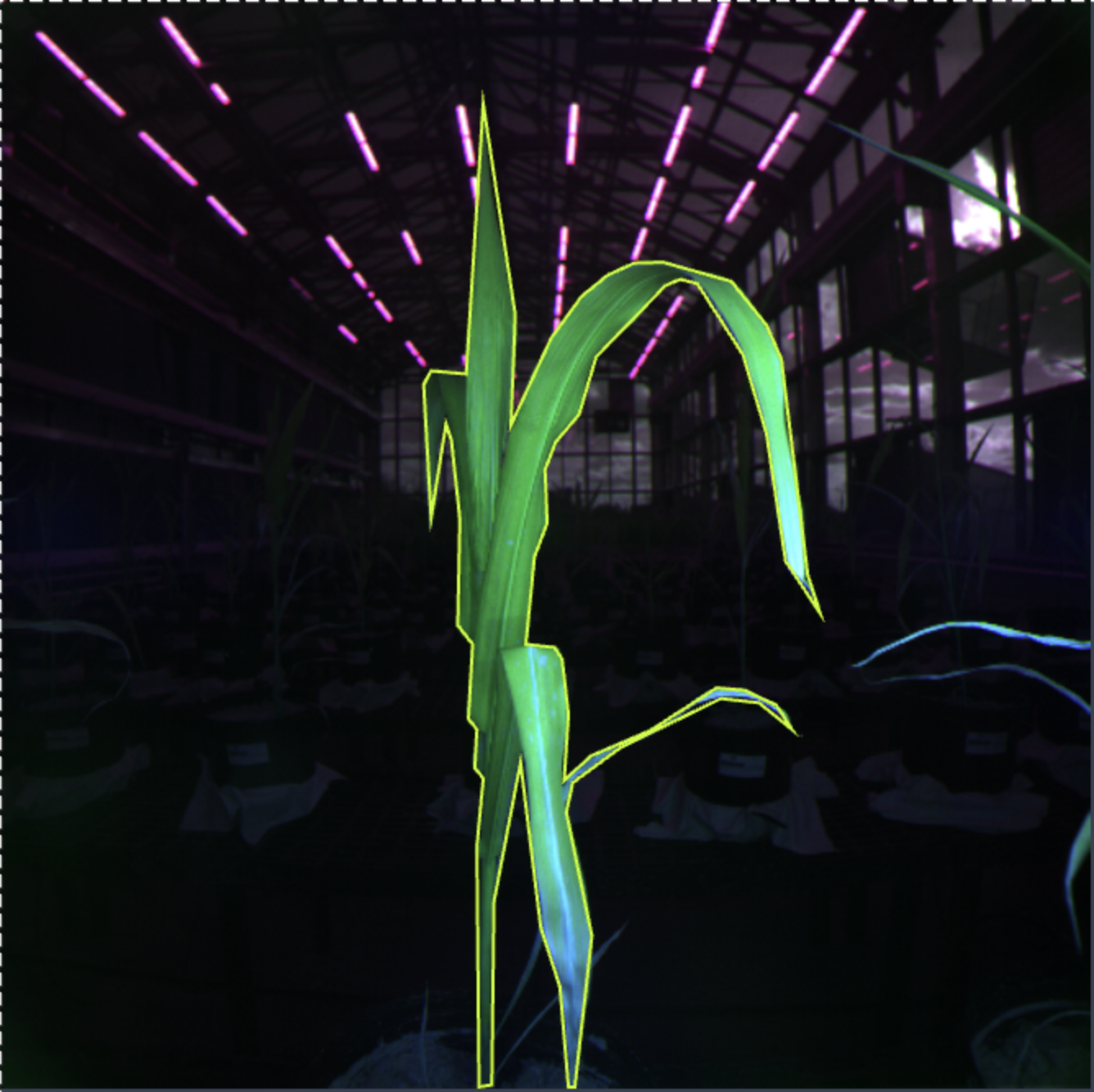}
        \caption{Corn Label}
        \label{fig:corn_label}
    \end{subfigure}\hfill % Adjust spacing between subfigures
    \begin{subfigure}[b]{0.3\linewidth}
        \centering
        \includegraphics[width=\linewidth]{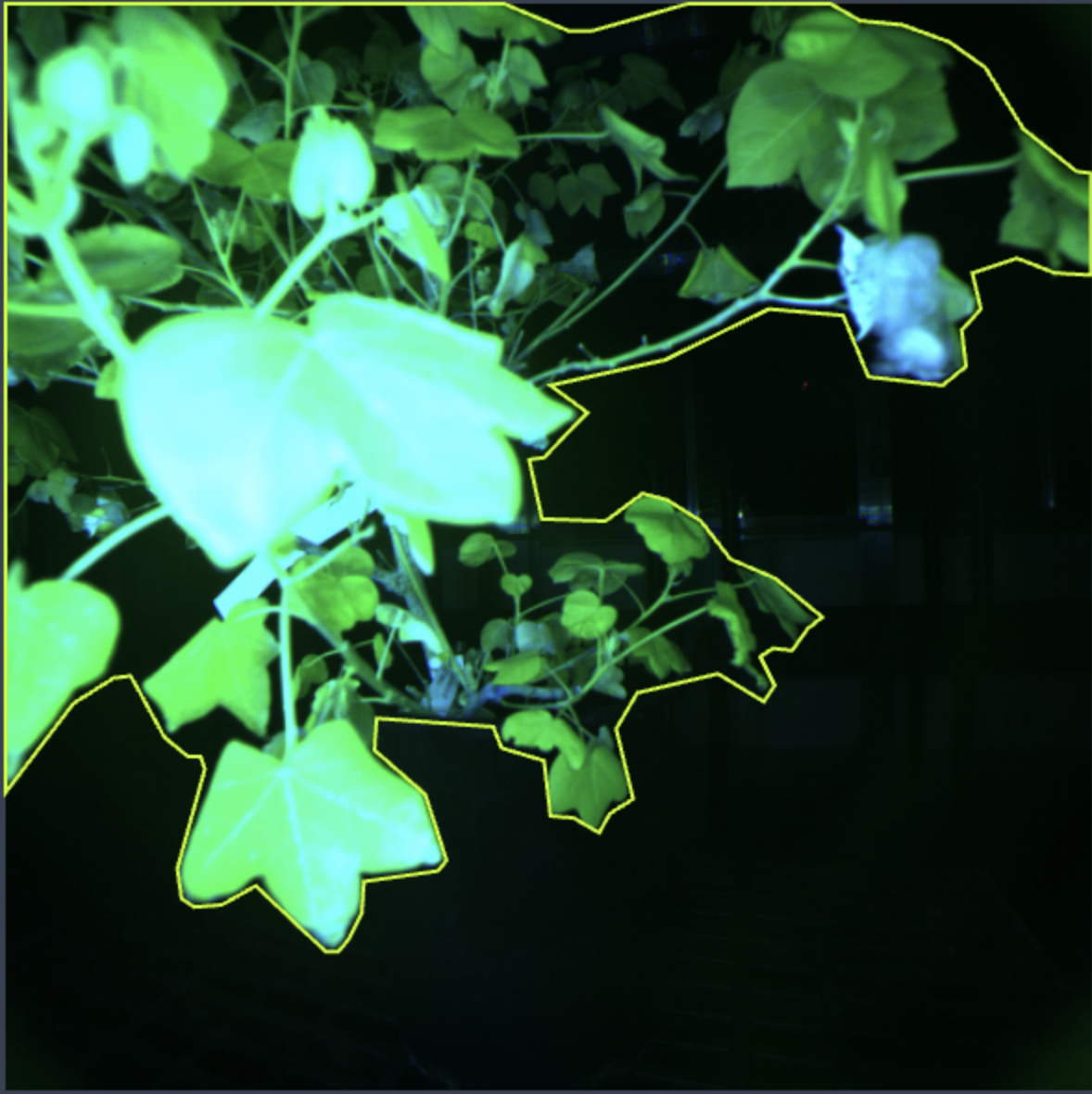}
        \caption{Cotton Label}
         \label{fig:cotton_label}
    \end{subfigure}\hfill
    \begin{subfigure}[b]{0.3\linewidth}
        \centering
        \includegraphics[width=\linewidth]{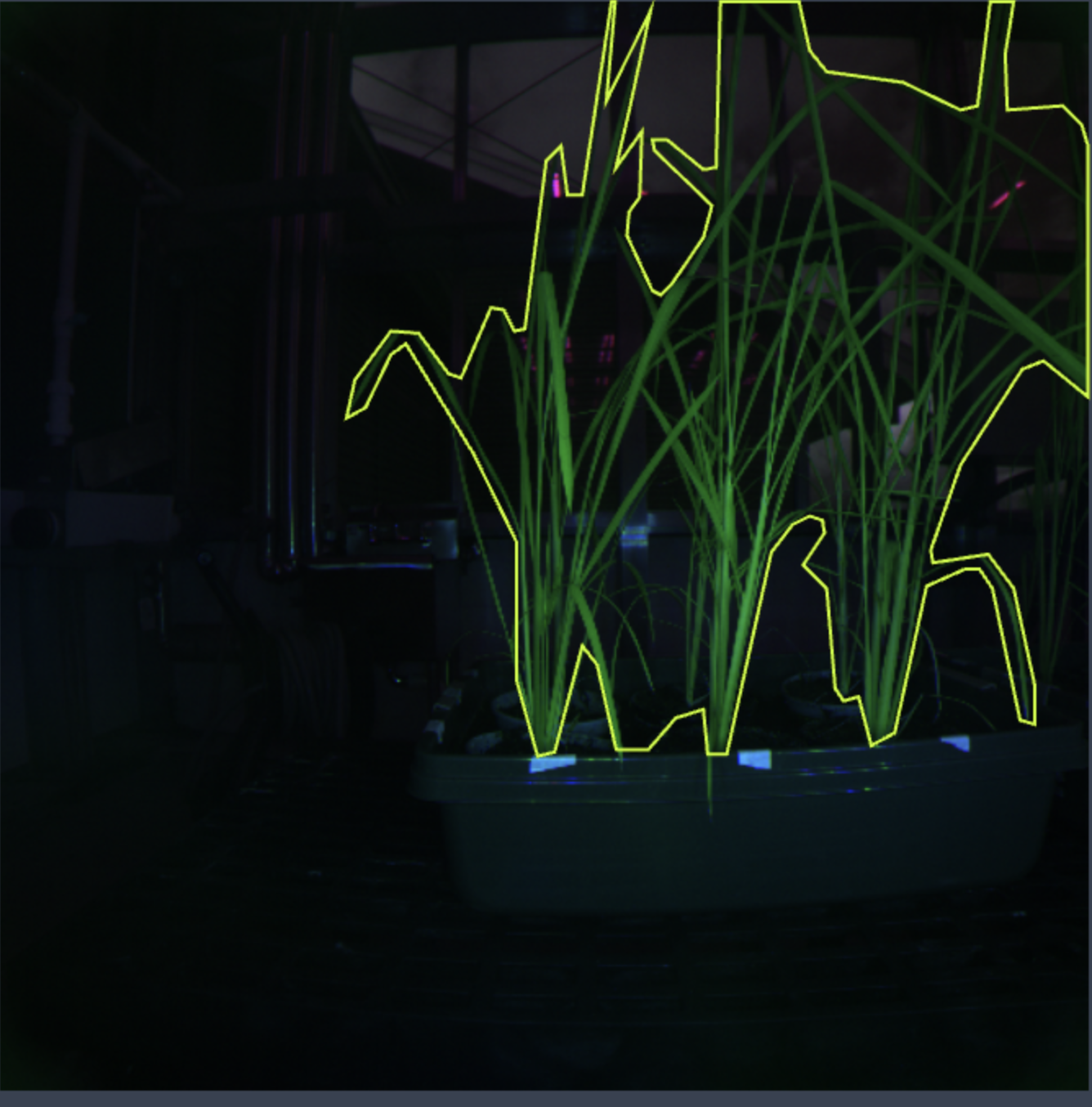}
        \caption{Rice Label}
         \label{fig:rice_label}
    \end{subfigure}
    \caption{Example images from PGP dataset. The top rows (\ref{fig:corn_img}-\ref{fig:rice_img}) are input images and the bottom rows (\ref{fig:cotton_label}-\ref{fig:rice_label}) are the associated labels.}
    \label{fig:PGPdataset}
\end{figure}
The Plant Growth and Phenotyping (PGP) dataset is a publicly available, high-resolution ($512 \times 512$ pixels), multi-spectral collection of plant images captured from a state-of-the-art greenhouse. PGP consists of 1137 images from different crops: 442 corn, 246 cotton, and 449 rice plants. The dataset includes images taken from various heights of the same plants. 
%and composite images (stitched images of the same plant). 
%Training a model with both multi-view and composite images ensures flexibility in object detection across various scales and conditions. While composite images excel at detecting complete plants, incorporating multi-view images enables the model to effectively detect plant parts in zoomed-in or cropped versions, ensuring robust performance regardless of the input image type.
\begin{figure*}[htb]
	\centering
	\includegraphics[width=.90\linewidth]{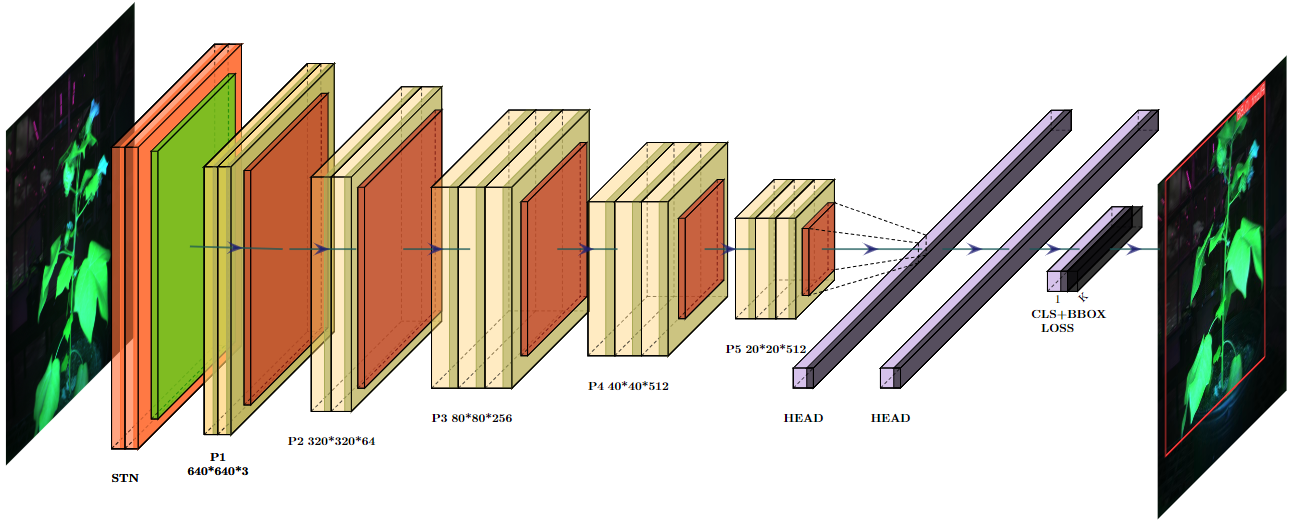}
	\caption{Architecture of STN-YOLO model is shown. The STN module (shown in light orange and green) is added at the beginning of the network to account for spatial invariances. After the STN, the image is passed into the YOLO model where the layers P1 - P5 are the layers of the YOLO backbone, HEAD part is responsible for generating the final output and the CLS + BBOX are losses to perform object detection.}
        \label{Architecture}
\end{figure*}
The plant phenotyping facility consisted of a camera mounted on a robotic arm, capturing side and top views of the plants (only side view images are used in this work). The Multispectral Imaging System (MSIS-AGRI-1-A), integrating an MSC2-AGRI-1-A snapshot multispectral camera and a 4-channel LED illuminator, was used to collect images across four spectral bands: 580 nm (Green), 660 nm (Red), 730 nm (Red Edge), and 820 nm (Near Infrared). 

For this work, the Red, Red Edge, and Green bands were combined to create pseudo ``RGB" images (as shown in Figure \ref{fig:PGPdataset}) compatible with pre-trained machine learning models. The individual bands were superimposed, and min-max normalization was applied to the final image. These plant images were captured between 2022 and 2024 in controlled greenhouse environments. Data augmentation techniques (e.g., rotation, cropping) can be applied using frameworks like Roboflow \cite{roboflow} to expand the dataset. The same framework is used for labeling, a two-step process involving auto-generation of initial labels using the Segment Anything Model (SAM) \cite{SAM}, followed by manual refining of the auto-generated labels.

\subsection{STN-YOLO MODEL}  \label{sect:STN-YOLO}  The STN module is first integrated in the beginning of the YOLO model to account for spatial transformations in the input images as shown in Figure \ref{Architecture} for the proposed STN-YOLO model. After the images passes through STN module, the YOLO model is then used to perform the task of object detection. As stated in Section \ref{sect:STN}, the STN has can use any backbone for the localization networks. The localization network within the STN module comprises various convolutional layers, linear layers, and activation functions.

\noindent\textbf{Localization in STN model} 
The STN models have the localization network which takes the input feature maps with width $W$ and height $H$ and $C$ channels and outputs $\theta$ which are the transformation parameters. The localization network is selected by the user and can vary from a shallow or deeper network. The performances of different localization network are compared to evaluate if a shallow or deep, pre-trained model will lead to better object detection performance quantitatively (\textit{e.g.}, accuracy, precision, and recall) and computationally. The shallow network is comprised of one convolution layer with max pooling and ReLU activation function was used in contrast to the deep pretrained models. Following the convolution layer in the shallow network, an adaptive average pooling layer was used to aggregate the spatial resolution of the feature maps to a desired size (\textit{e.g.}, $28 \times 28$).

\noindent\textbf{YOLO Backbone} The output of the STN module is then given to the YOLO model where, P1, P2, P3, P4, and P5 typically refer to different levels or stages in the feature pyramid and these are the different convolutional layers as shown in the Figure \ref{Architecture}. Multi-scale information is obtained by extracting features at different scales from the input image  to detect objects of various sizes and capture both fine-grained details and global context. P1 corresponds to the finest level with the highest resolution, while P5 is the coarsest level with lower resolution but capturing more context. The head in the YOLO model usually refers to the part of the network responsible for making predictions. The head takes features from multiple pyramid levels and produces predictions for object classes, bounding box coordinates, and other relevant information. 

\noindent\textbf{Loss in YOLOv8}
A two term loss function is used during training. The objective function consists of two components: classification loss (CLS), which penalizes errors in predicting object classes, and bounding box (BBOX) loss, which penalizes errors in predicting the coordinates of the bounding boxes around objects. Combining these losses helps train the model to simultaneously improve both classification and localization aspects of object detection. Class imbalance in single-class detection refers to the discrepancy between detection target objects and the background, typically considered the negative class. Addressing this imbalance is crucial for improving detection performance. Bounding box loss in YOLOv8 uses the complete intersection over union (CIOU) loss \cite{CIOU}, which considers box overlap and aspect ratio differences, enhancing regression performance. Distributed focal loss (DFL) \cite{DFL} effectively tackles class discrepancy by dynamically adjusting the loss for each class during training to counteract biases caused by class imbalances.

\begin{table*}[htb]
    \centering
    \caption[Localization performance metrics]{Performance metrics for different localization networks with the average value and +/- 1 standard deviation are shown across three experimental runs of 100 epochs each with random initialization. The best average metric is bolded.}
    \begin{tabular}{|c|c|c|c|c|}
        \hline
        Localization & Accuracy & Precision & Recall & mAP \\
        \hline
        Shallow  & 80.34 $\pm$ 0.92 & \textbf{94.33 $\pm$ 0.74} & 87.62 $\pm$ 0.71 & 71.82 $\pm$ 0.58 \\
        \hline
        VGG16 & \textbf{81.05 $\pm$ 0.08} & 92.94 $\pm$ 0.39 & \textbf{88.50 $\pm$ 0.92} & \textbf{72.87 $\pm$ 0.66} \\
        \hline
        Resnet18  & 80.97 $\pm$ 0.74 & 91.51 $\pm$ 0.09 & 88.03 $\pm$ 0.69 & 72.04 $\pm$ 0.71 \\
        \hline
    \end{tabular}
    
    \label{tab:localization_metrics}
\end{table*}

\begin{table*}[htb]
    \centering
    \caption[Performance metrics for different spatial resolutions]{Metrics for different spatial resolutions of the STN shallow localization network with  the average value and +/- 1 standard deviation are shown across three experimental runs of 100 epochs each with random initialization. The best average metric is bolded.}
    \begin{tabular}{|c|c|c|c|c|}
        \hline
        Feature Map Size & Accuracy & Precision & Recall & mAP \\
        \hline
         $1 \times 1$  & 80.34 $\pm$ 0.92 & 94.33 $\pm$ 0.74 & 87.62 $\pm$ 0.71 & 71.82 $\pm$ 0.58 \\
        \hline
         $7 \times 7$ &  80.73 $\pm$ 0.57 & 95.17 $\pm$ 0.80 & 88.71 $\pm$ 0.46 & 71.39 $\pm$ 0.82 \\
        \hline
         $28 \times 28$ & \textbf{81.63 $\pm$ 1.53} & \textbf{95.34} $\pm$ \textbf{0.76} & \textbf{89.52} $\pm$ \textbf{0.57} & \textbf{72.56} $\pm$ \textbf{0.90} \\
        \hline
    \end{tabular}
    
    \label{tab:adaptive_metrics}
\end{table*}

\section{Experimental Results and Discussion}

\subsection{Experimental Setup}
 Three experimental runs of random initialization were used for each model and the YOLO backbones were pretrained on the COCO dataset. We used the AdamW optimizer with an initial learning rate of 0.002. The optimizer was configured with distinct parameter groups, each with specific weight decay values assigned to all parameters. The batch size for all experiments was set to 16. We used 100 epochs with early stopping (patience of 50) for each model. All the experiments were performed using the Ultralytics \cite{yolov8_ultralytics} framework and performed on an NVIDIA A100 GPU.

In our study, we used various benchmark datasets to assess the effectiveness of our proposed plant detection models.
The GlobalWheat2020 dataset contains 4,000 images, with 3,000 for training and 1,000 for testing, focusing on wheat detection.
The PlantDoc dataset comprises 2,569 images, with 2,330 for training and 239 for testing, addressing plant identification across varying environments.
The MelonFlower dataset includes 288 images, with 193 for training and 95 for testing, specifically targeting melon flower identification. These datasets collectively provide a robust evaluation framework for assessing model generalization across diverse plant species and environments.

\subsection{Ablation Studies using PGP Dataset}

\noindent\textbf{Impact of Localization Networks}
In Table \ref{tab:localization_metrics}, we outline the performance metrics of various localization networks, emphasizing the comparable quantitative results achieved by both the shallow localization network and deep pre-trained models.  Both the shallow and deep networks used a global average pooling layer after the convolutional layers. Based on these findings, we opted for the shallow network as the localization network in the STN-YOLO model. This decision stemmed from the observed effectiveness of the shallow network in extracting features suitable for learning affine transformations comparable to the capabilities of the pre-trained models (based on similar quantitative measures). The shallow network and the pretrained networks were finetuned  with the baseline YOLO model. The number of learnable parameters for the deeper networks, VGG16 and ResNet18 (approximately 82 million and 49 million respectively), were greater than the shallow network (approximately 5 million) for marginal improvements in performance hence the shallow network was chosen for the remaining experiments.

% The YOLO model with the simple localization network boasts 5,821,078 parameters, where as the ResNet increases this number to 49,426,806, and VGG localization dramatically escalates it to 82,007,110, indicating substantial performance enhancements or complexity with each subsequent model. However, the increase in performance was marginal, leading to the decision to prioritize computational efficiency. It's noteworthy that while the shallow network underwent training on our dataset, the pre-trained models remained fixed. This decision was driven by computational constraints, as we fine tuned a few layers of pretrained models for VGG16 and RESNET18 but this just eventually increased the computation cost and the performance was not significantly greater. Hence to save on hardware and computatations the decision to use the shallow network was taken.

\noindent\textbf{Impact of Spatial Information}
To assess the impact of retaining more spatial information in the STN-YOLO model, modifications were applied to the last layer of the shallow localization network. Specifically, the adaptive average pooling layer, was evaluated for various spatial sizes of $1 \times 1$, $7 \times 7$, and $28 \times 28$ as shown in Table \ref{tab:adaptive_metrics}. This investigation aims to evaluate the model's sensitivity to variations in spatial information preservation and identify the optimal configuration for improved performance in object detection tasks. 
We wanted to look at no spatial information ($1 \times 1$) and retaining some spatial information ($7 \times 7$, $28 \times 28$). The results demonstrated that retaining the most amount of spatial information ($28 \times 28$) resulted in the best object detection performance. If the feature maps have more spatial information, then there is more context that the model will be able to use for maximizing performance. However, as the spatial resolution is increased, so does the computational cost as the number of parameters will increase. 

\begin{figure*}[htb]
    \centering
    \setlength{\fboxsep}{0pt} % Remove padding around subfigures
    \begin{subfigure}[b]{0.17\linewidth} % Adjust width of subfigures
        \centering
        \includegraphics[width=\linewidth]{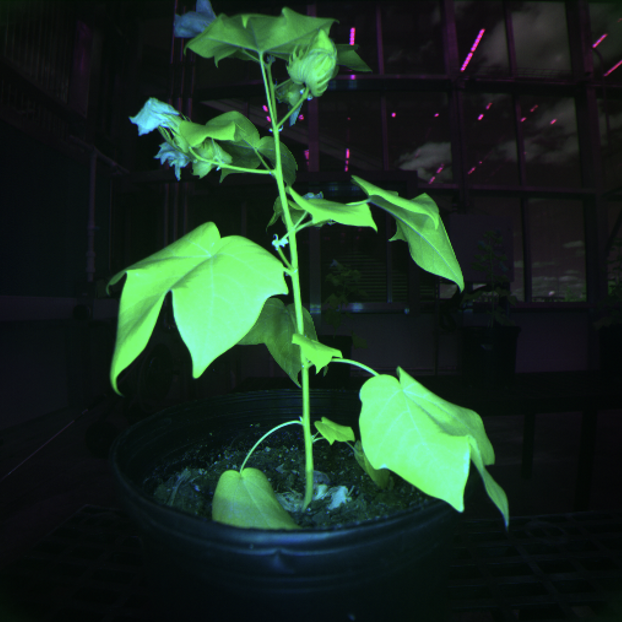}
        \caption{Cotton}
        \label{fig:cotton_img_2}
    \end{subfigure}\hfill % Adjust spacing between subfigures
    \begin{subfigure}[b]{0.17\linewidth}
        \centering
        \includegraphics[width=\linewidth]{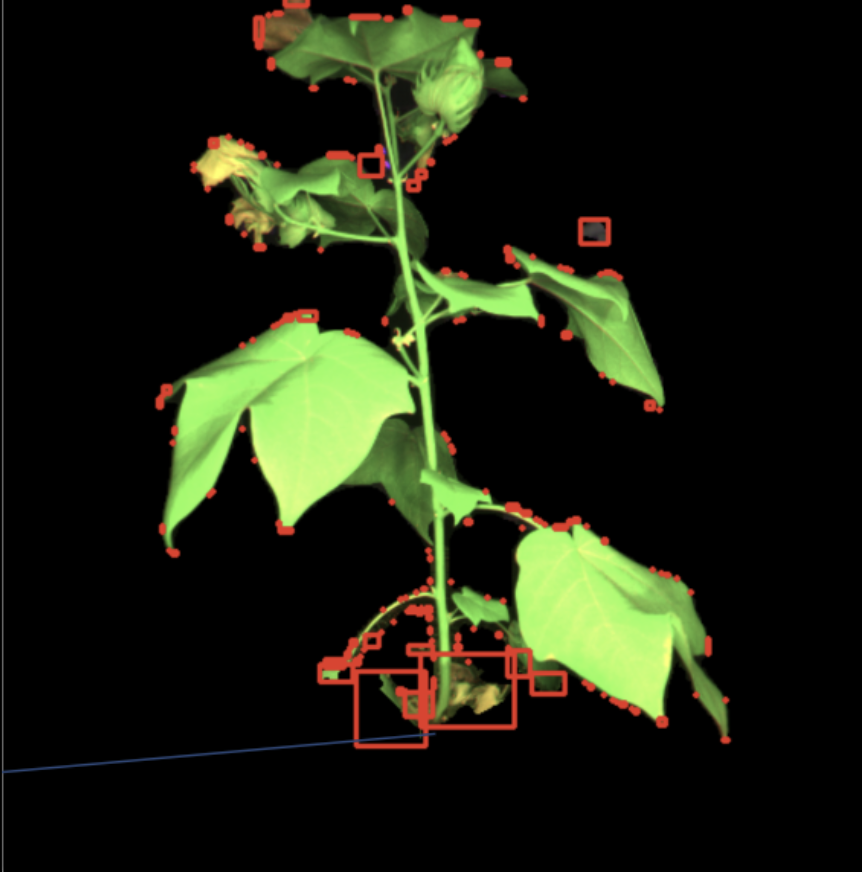}
        \caption{YOLO}
        \label{yolodetection}
    \end{subfigure}\hfill
    \begin{subfigure}[b]{0.17\linewidth}
        \centering
        \includegraphics[width=\linewidth]{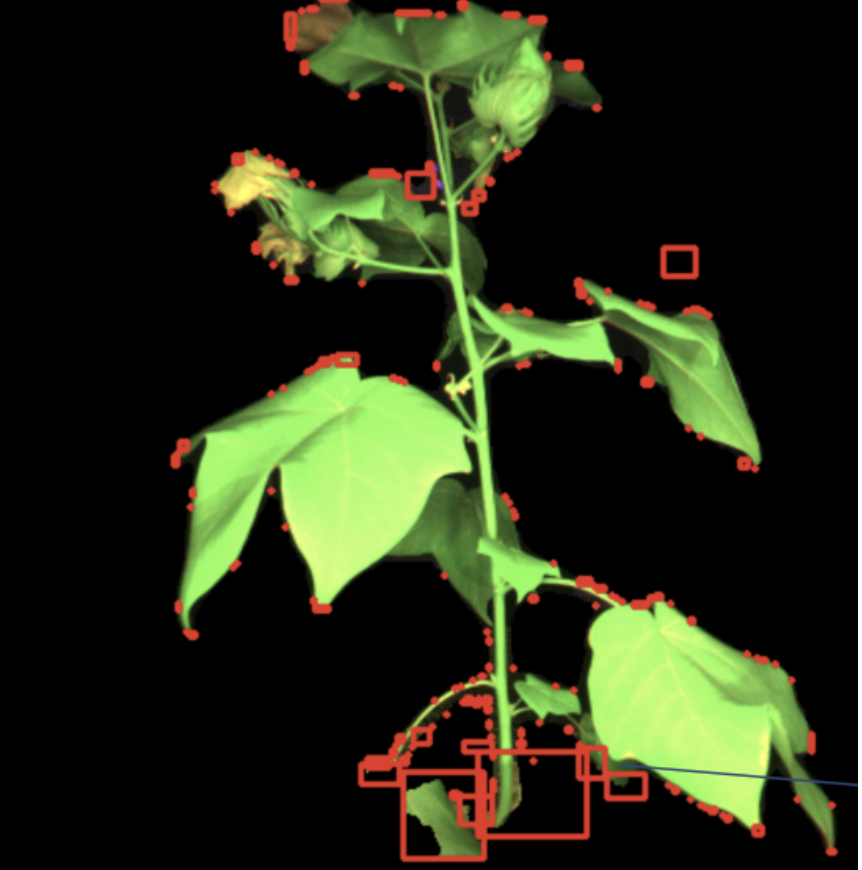}
        \caption{STN-YOLO (ours)}
        \label{stndetection}
    \end{subfigure}\hfill
    \begin{subfigure}[b]{0.17\linewidth} % Adjust width of subfigures
        \centering
        \includegraphics[width=\linewidth]{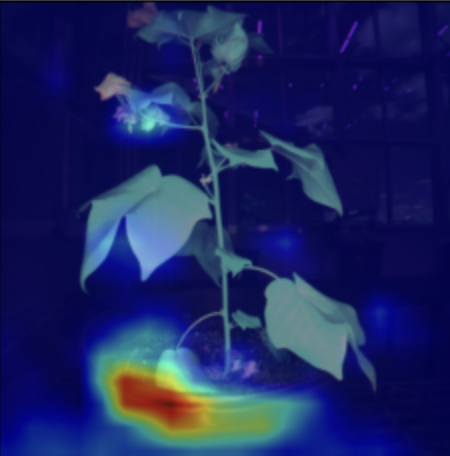}
        \caption{YOLO}
        \label{classyolo}
    \end{subfigure}\hfill % Adjust spacing between subfigures
    \begin{subfigure}[b]{0.17\linewidth}
        \centering
        \includegraphics[width=\linewidth]{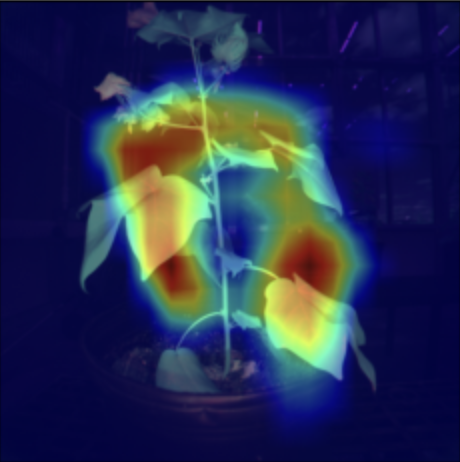}
        \caption{STN-YOLO (ours)}
        \label{classSTN-YOLO}
    \end{subfigure}\hfill
\caption{Example results of YOLO and STN-YOLO on an example image from PGP dataset (Figure \ref{fig:cotton_img_2}). We show the detection differences in the YOLO and the STN-YOLO models in Figure \ref{yolodetection} and \ref{stndetection}. The YOLO model does not capture portions of the leaf in the pot and falsely detects a portion of the background. EigenCAM \cite{muhammad2020eigen} was used to show which areas of the image each model focused on for object detection. STN-YOLO focused on the area of the image containing a majority of the plant as opposed to YOLO emphasizing the pot of the plant.}
\end{figure*}

\noindent\textbf{Model Analysis using Explainable AI} The proposed model is better on average at reducing the detection of false positives, as the precision of the STN-YOLO model is better than the baseline YOLO model as shown in Table \ref{tab:new_augmentation_results}. The STN-YOLO model also exhibits improved recall and mean average precision (mAP), which demonstrates a better object detection model as illustrated in Figure (\ref{yolodetection}-\ref{stndetection}), where the baseline YOLO model fails to eliminate the bottom right vase area in comparison to the proposed STN-YOLO model. In addition to these metrics from Table \ref{tab:new_augmentation_results}, we also provide insights from Eigen class activation maps (EignenCAMs) \cite{muhammad2020eigen} as seen in Figure (\ref{classyolo}-\ref{classSTN-YOLO}). These EigenCAMs offer valuable visualization for model explainability, as they highlight significant aspects of the images that contribute to the model's predictions. The results support the hypothesis that incorporating the spatial invariance in the input image can result in better object detection both qualitatively and quantitatively as the STN-YOLO emphasizes more of the plant area.

\begin{table*}[htb]
    \centering
    \caption[Object detection metrics Comparison for different augmentations]{Object detection metrics for different augmentations individually and together with the average value and +/- 1 standard deviation are shown across three experimental runs of 100 epochs each with random initialization. The best average metric is bolded.}
    \begin{tabular}{|c|c|c|c|c|c|c|c|}
    \hline
    Rotation & Shear & Crop & Model & Accuracy & Precision & Recall & mAP \\ \hline
     & & & YOLO  & \textbf{84.86 $\pm$ 0.47} & 94.30 $\pm$ 0.56 & 89.21 $\pm$ 0.53 & 71.76 $\pm$ 1.03 \\ \cline{4-8}
     & & &  STN-YOLO  & 81.63 $\pm$ 1.53 & \textbf{95.34} $\pm$ \textbf{0.76} & \textbf{89.52} $\pm$ \textbf{0.57} & \textbf{72.56} $\pm$ \textbf{0.90}\\ \hline
    &  & \checkmark & YOLO & 83.81 $\pm$ 2.29  & \textbf{94.87 $\pm$ 0.87} & 87.90 $\pm$ 0.84 & \textbf{71.26 $\pm$ 1.56} \\ \cline{4-8}
     & & & STN-YOLO & \textbf{84.52 $\pm$ 2.37} & 94.58 $\pm$ 1.01 & \textbf{88.44 $\pm$ 0.74} & 70.06 $\pm$ 0.64 \\ \hline
    & \checkmark &  & YOLO & \textbf{84.86 $\pm$ 0.97}  & 93.06 $\pm$ 1.23 & 88.70 $\pm$ 0.30 & \textbf{72.35 $\pm$ 1.22} \\ \cline{4-8}
     & & & STN-YOLO & 83.38 $\pm$ 1.61 & \textbf{94.87} $\pm$ \textbf{0.85} & \textbf{89.64 $\pm$ 0.54} & 68.10 $\pm$ 1.56 \\ \hline
     & \checkmark & \checkmark &  YOLO & 82.73 $\pm$ 2.60 & 93.94 $\pm$ 0.31 & \textbf{88.96 $\pm$ 0.70} & \textbf{71.97 $\pm$ 1.08} \\ \cline{4-8}
     & & & STN-YOLO & \textbf{83.36 $\pm$ 0.55} & \textbf{94.58} $\pm$ \textbf{0.82} & 87.67 $\pm$ 0.20 & 69.09 $\pm$ 0.28 \\ \hline
    \checkmark &  &  & YOLO & 83.46 $\pm$ 0.71 & 94.87 $\pm$ 0.82 & 88.71 $\pm$ 0.31 & 70.65 $\pm$ 0.68 \\ \cline{4-8}
     & & & STN-YOLO & \textbf{85.39 $\pm$ 0.69} & \textbf{95.47} $\pm$ \textbf{1.38} & \textbf{89.91 $\pm$ 0.72} & \textbf{71.54 $\pm$ 0.43}\\ \hline
     \checkmark & & \checkmark & YOLO & \textbf{85.43 $\pm$ 0.48}  & \textbf{95.09} $\pm$ \textbf{1.61} & \textbf{89.91 $\pm$ 0.74} & \textbf{70.67 $\pm$ 0.57} \\ \cline{4-8}
     & & & STN-YOLO & 82.66 $\pm$ 1.67  & 94.54 $\pm$ 1.12 & 88.09 $\pm$ 0.19 & 68.69 $\pm$ 0.63 \\ \hline
    \checkmark & \checkmark & & YOLO & \textbf{88.59 $\pm$ 2.04} & 94.87 $\pm$ 0.84 & 89.02 $\pm$ 0.61 & \textbf{72.97 $\pm$ 1.12} \\ \cline{4-8}
     & & & STN-YOLO & 82.78 $\pm$ 1.46 & \textbf{96.54} $\pm$ \textbf{1.52} & \textbf{89.02 $\pm$ 0.69} & 71.31 $\pm$ 0.61 \\ \hline
    \checkmark & \checkmark & \checkmark & YOLO & \textbf{84.86 $\pm$ 0.91} & 93.02 $\pm$ 0.43 & 88.49 $\pm$ 0.31 & \textbf{72.35 $\pm$ 1.46} \\ \cline{4-8}
     & & & STN-YOLO & 84.60 $\pm$ 1.28 & \textbf{94.62} $\pm$ \textbf{1.84} & \textbf{89.53 $\pm$ 0.65} & 69.83 $\pm$ 0.73 \\ \hline
    \end{tabular}
    \label{tab:new_augmentation_results}
\end{table*}

\begin{table*}[htb]
    \centering
    \caption[Metrics for Benchmark Datasets]{Object detection performance metrics for each model on benchmark Agricultural datasets with the average value and +/- 1 standard deviation are shown across three experimental runs of 100 epochs each with random initialization. The best average metric is bolded for each dataset.}

\centering

\begin{tabular}{|c|c|c|c|c|c|}
\hline
Datasets                           & Model         & Accuracy                 & Precision                 & Recall                    & mAP                        \\ \hline
 PGP                   & YOLO  & \textbf{84.86 $\pm$ 0.47} & 94.30 $\pm$ 0.56 & 89.21 $\pm$ 0.53 & 71.76 $\pm$ 1.03 \\ \cline{2-6} 
   & STN-YOLO  & 81.63 $\pm$ 1.53 & \textbf{95.34} $\pm$ \textbf{0.76} & \textbf{89.52} $\pm$ \textbf{0.57} & \textbf{72.56} $\pm$ \textbf{0.90}\\ \hline
Global Wheat 2020                  & YOLO & \textbf{94.29 $\pm$ 0.10} & 92.17 $\pm$ 0.03        & 96.69 $\pm$ 0.29         & 62.67 $\pm$ 0.25         \\ \cline{2-6} 
                                   & STN-YOLO      & 93.67 $\pm$ 0.79         & \textbf{92.24 $\pm$ 0.24}        & \textbf{97.53 $\pm$ 0.02}         & \textbf{63.32 $\pm$ 0.71}         \\ \hline
PlantDoc                           & YOLO & 47.57 $\pm$ 0.40         & 47.21 $\pm$ 0.15        & 47.13 $\pm$ 1.19         & 30.29 $\pm$ 0.84        \\ \cline{2-6} 
                                   & STN-YOLO      & \textbf{49.57 $\pm$ 0.74}         & \textbf{52.01 $\pm$ 2.76} & \textbf{48.47 $\pm$ 1.16} & \textbf{34.79 $\pm$ 2.02} \\ \hline
MelonFlower                       & YOLO & 86.47 $\pm$ 2.37         & 73.63 $\pm$ 2.83        & 85.27 $\pm$ 0.82         & 44.91 $\pm$ 0.47         \\ \cline{2-6} 
                                   & STN-YOLO      & \textbf{89.78 $\pm$ 1.15} & \textbf{88.09 $\pm$ 3.44}         & \textbf{86.97 $\pm$ 0.72}         & \textbf{45.09 $\pm$ 3.31}         \\ \hline
\end{tabular}
\label{tab:Benchmark Datasets Results}
\end{table*}

\noindent\textbf{Augmentations Testing} One of the most important set of experiments performed was the testing  of models on the unseen augmented data as shown in Table \ref{tab:new_augmentation_results}. To do this, no data augmentation was added to the training dataset, but data augmentations were performed on the test dataset to evaluate the robustness of the STN-YOLO model across various transformations. The augmentations used on the test dataset were random cropping (15\% zoom), shear ($\pm$ $10^\circ$ horizontal and $\pm$ $10^\circ$ vertical), and rotation (between $\pm$ $10^\circ$) to simulate image conditions that may happen in the facility. We observed that the proposed STN-YOLO model gave better quantitative results in case of rotation and shear for PGP dataset compared to the baseline YOLO model. This result is intuitive as rotation and shear are affine transformations that can be learned by the STN-YOLO model. However, in the cases where cropping is used, the performance is comparable to the baseline YOLO method as cropping is not an affine transformation. Furthermore, for the combinations of the different augmentations (\textit{e.g.}, rotation with shear, rotation with crop), we observed that for most of the data augmentation cases (six out of eight), the STN-YOLO model performs better on average than the YOLO baseline for the precision values indicating that the STN-YOLO was effective at reducing the number of false positives by incorporating the added spatial invariance properties of the STNs.

\begin{figure}[htb]
    \centering
    \setlength{\fboxsep}{0pt} % Remove padding around subfigures
    
    % Title for the first row of images
    \begin{minipage}{\textwidth}
        {\hspace{0.17cm} Ground Truth \hspace{1.4cm} YOLO \hspace{1.55cm} STN-YOLO}
    \end{minipage}
    
    \begin{subfigure}[b]{0.29\linewidth} % Adjust width of subfigures
        \centering
        \includegraphics[width=1\linewidth]{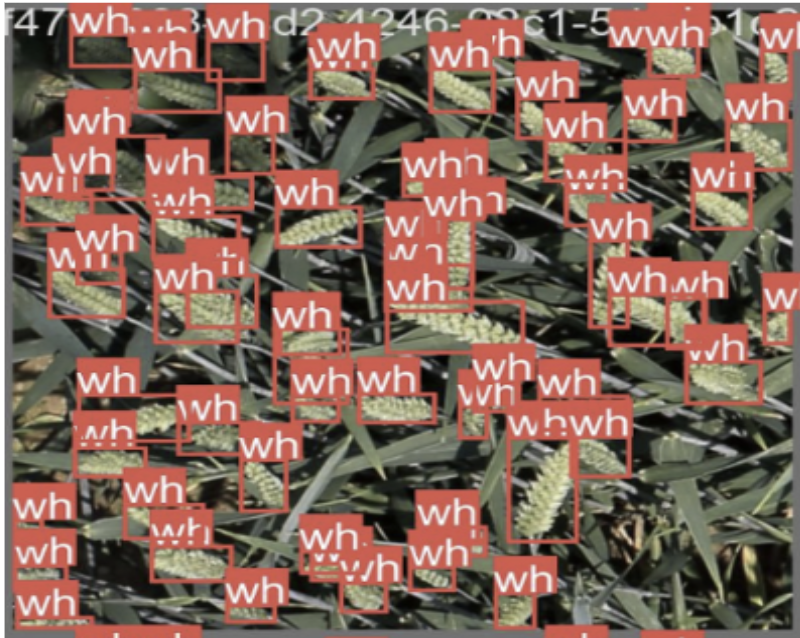}
        
        \label{fig:wheat_label}
    \end{subfigure}\hfill % Adjust spacing between subfigures
    \begin{subfigure}[b]{0.3\linewidth}
        \centering
        \includegraphics[width=\linewidth]{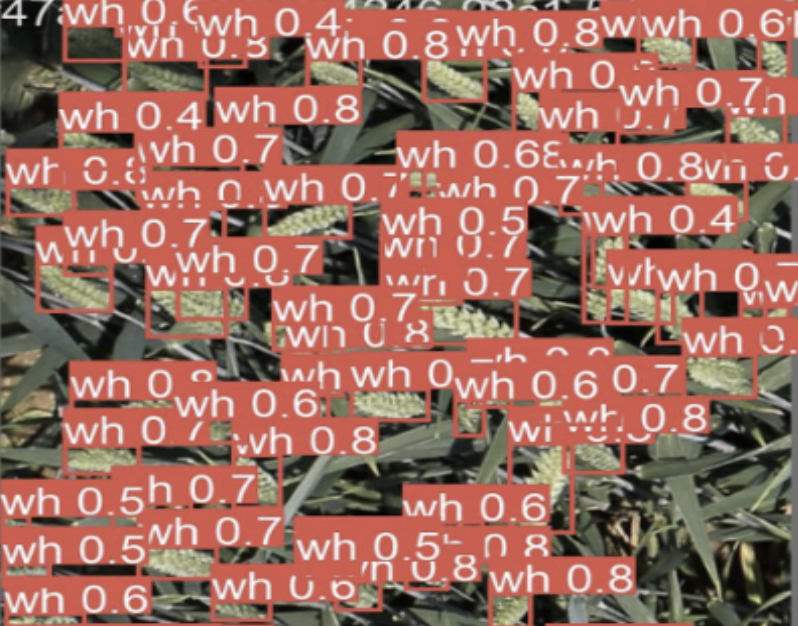}
        
        \label{fig:wheat_yolo}
    \end{subfigure}\hfill
    \begin{subfigure}[b]{0.3\linewidth}
        \centering
        \includegraphics[width=\linewidth]{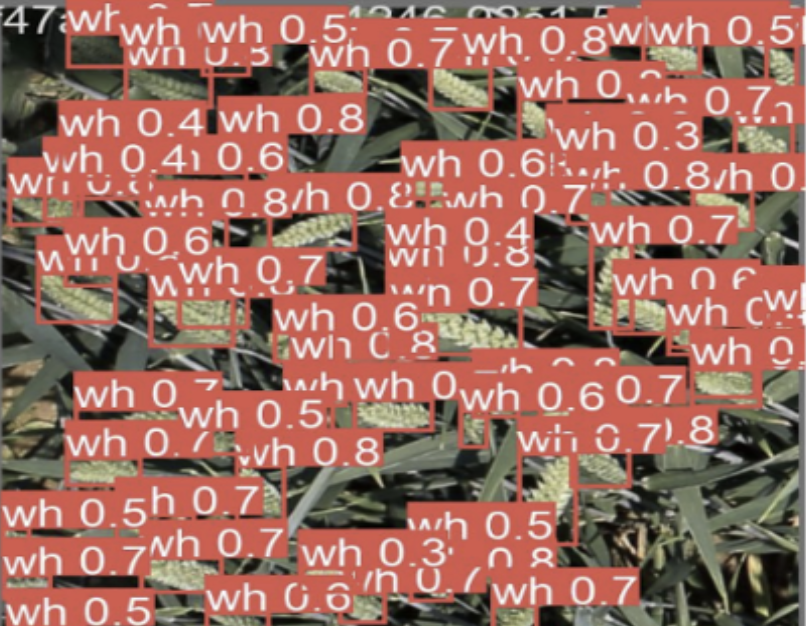}
        
        \label{fig:wheat_stn}
    \end{subfigure}
    
    % Subcaption for the first row
    \begin{minipage}{0.3\textwidth}
    
    \centering
    \subcaption{GlobalWheat2020}\label{subfig:globalwheat}
    \end{minipage}

    \begin{subfigure}[b]{0.3\linewidth} % Adjust width of subfigures
        \centering
        \includegraphics[width=1\linewidth]{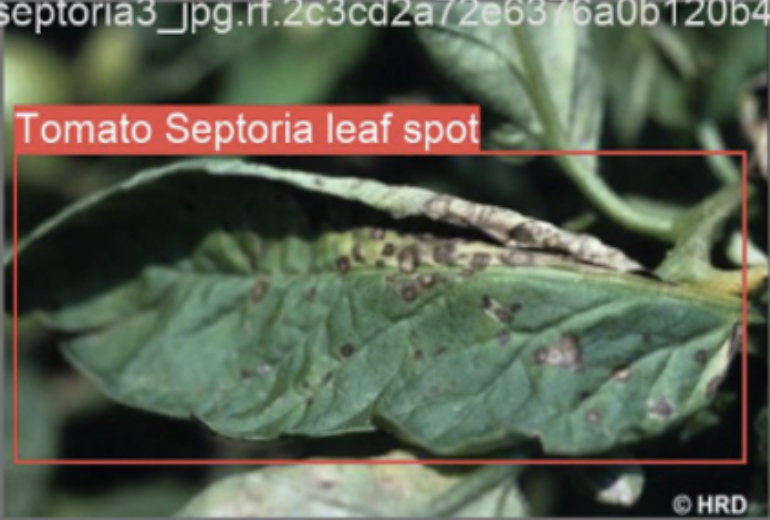}
        
        \label{fig:pd_label}
    \end{subfigure}\hfill % Adjust spacing between subfigures
    \begin{subfigure}[b]{0.3\linewidth}
        \centering
        \includegraphics[width=\linewidth]{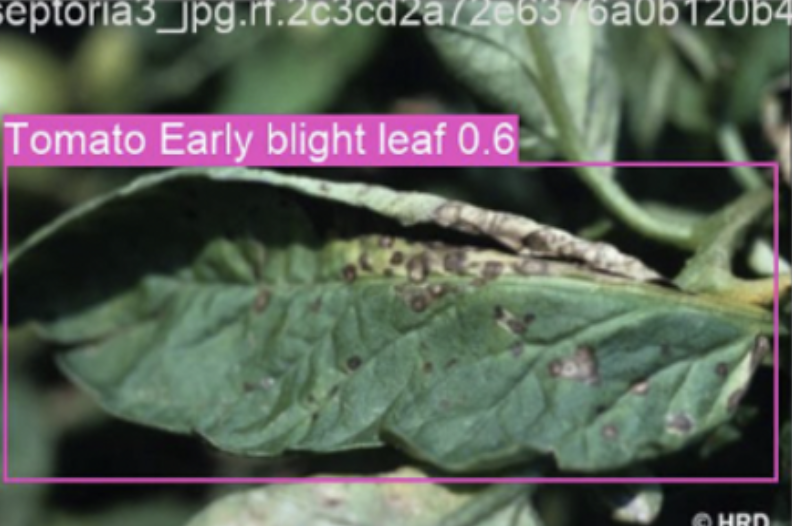}
        
         \label{fig:pd_yolo}
    \end{subfigure}\hfill
    \begin{subfigure}[b]{0.3\linewidth}
        \centering
        \includegraphics[width=\linewidth]{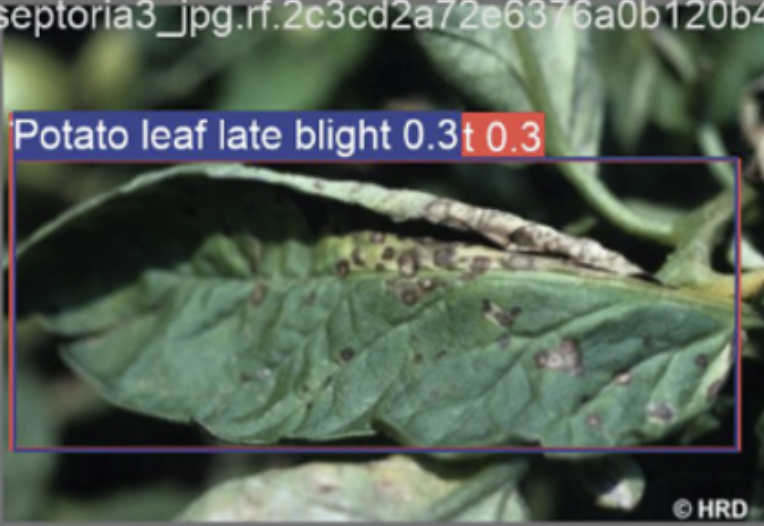}
        
         \label{fig:pd_stn}
    \end{subfigure}
    
    % Subcaption for the second row
    \begin{minipage}{0.3\textwidth}
    
    \subcaption{PlantDoc}\label{subfig:plantdoc}
    \end{minipage}

    \begin{subfigure}[b]{0.3\linewidth} % Adjust width of subfigures
        \centering
        \includegraphics[width=1\linewidth]{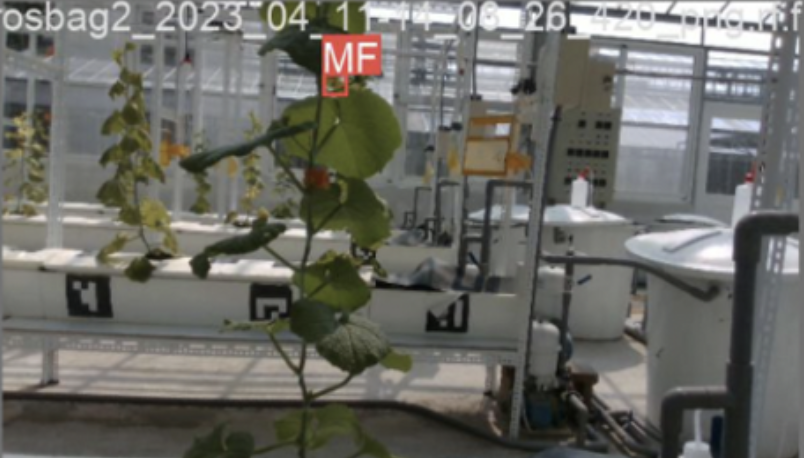}
        
        \label{fig:mf_label}
    \end{subfigure}\hfill % Adjust spacing between subfigures
    \begin{subfigure}[b]{0.3\linewidth}
        \centering
        \includegraphics[width=\linewidth]{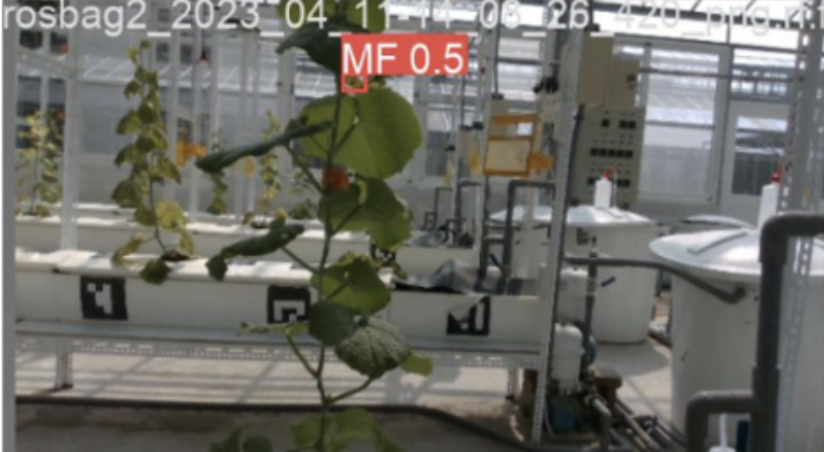}
        
         \label{fig:mf_yolo}
    \end{subfigure}\hfill
    \begin{subfigure}[b]{0.3\linewidth}
        \centering
        \includegraphics[width=\linewidth]{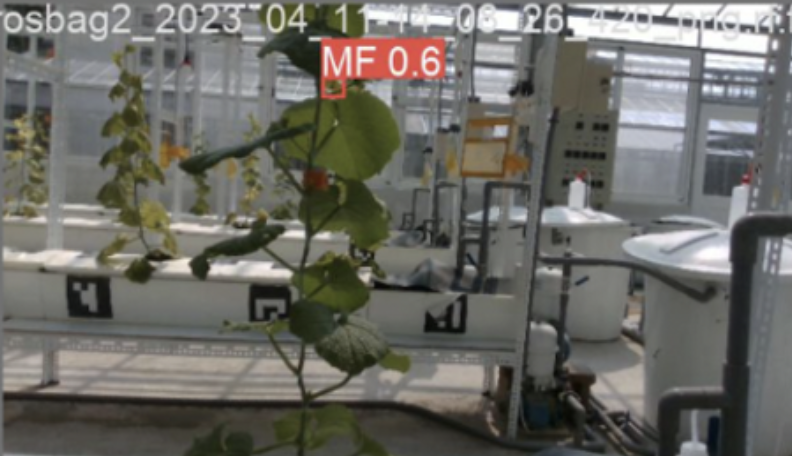}
        
         \label{fig:mf_stn}
    \end{subfigure}
    
    % Subcaption for the third row
    \begin{minipage}{0.3\textwidth}
    
    \subcaption{MelonFlower}\label{subfig:melonflower}
    \end{minipage}
    
    % Subcaptions below the figure
    \caption{Example images from benchmark dataset. The top row in Figure \ref{subfig:globalwheat} are the GlobalWheat2020 dataset illustrations with labels followed the YOLO and STN-YOLO detections.  Figure \ref{subfig:plantdoc} are associated with the PlantDoc dataset and  Figure \ref{subfig:melonflower} are the MelonFlower dataset results for the labels, YOLO and the STN-YOLO detections.}
    \label{fig:benchmark_datasets}
    
\end{figure}

\subsection{Benchmark Datasets Results}

\noindent\textbf{GlobalWheat2020 Dataset} The GlobalWheat2020 dataset \cite{gw2020} is a comprehensive Agricultural dataset containing images of wheat heads sourced from various countries. From the results in Table \ref{tab:Benchmark Datasets Results}, the  STN-YOLO model has a higher average value for most metrics (except accuracy), but the baseline YOLO model is comparable to the average value when you take the standard deviation in comparison. We can also qualitatively look at this observation from Figure \ref{subfig:globalwheat} where the number of detections by YOLO in the GlobalWheat2020 dataset are less as compared to the detections of STN-YOLO.
For the GlobalWheat2020 dataset, the images consisted of ``zoomed"-in views of the wheatheads. In this case, the spatial invariance would not impact the object detection results and this is indicated by the slight improvements in the average object detection metrics (except for accuracy) shown in Table \ref{tab:Benchmark Datasets Results}.

\noindent\textbf{PlantDoc Dataset} The PlantDoc dataset \cite{singh2019plantdoc} is derived from the Plant Village dataset (PVD) \cite{plantvillage_dataset}. The dataset was created using an automated system leveraging GoogleNet \cite{googlenet} and AlexNet \cite{alexnet}. Interesting observations can also be seen from the Table \ref{tab:Benchmark Datasets Results} such as unlike the GlobalWheat2020 dataset, there is an increase in all the object detection metrics in the STN-YOLO model as compared to the baseline model. Particularly, the precision for STN-YOLO model is higher than the baseline model indicating the model reduced the number of false positives detected as noted for the PGP dataset. Qualitatively, we can see that in Figure \ref{subfig:plantdoc} where the YOLO detects the leaf correctly but wrongly classifies the leaf as tomato early blight leaf. However, the STN-YOLO detects and correctly classifies the leaf as Tomato Septoria. In the PlantDoc dataset, the emphasis remains on various plants and zoomed-in images. However, spatial invariance remains insignificant due to the absence of pre-translated or rotated input images. Despite this, spatial invariance could potentially offer some benefits with this dataset, particularly as certain images provide a more ``zoomed-out" perspective of the plants compared to the GlobalWheat2020 dataset.

\noindent\textbf{MelonFlower Dataset} The MelonFlower dataset \cite{melon_flower_dataset} is dataset available on Roboflow and the images were captured in a greenhouse (similar to the PGP dataset). This dataset was used to demonstrate the model's capability in detecting small objects (\textit{e.g.}, flowers) and different viewing geometries of the plants. As seen from Table \ref{tab:Benchmark Datasets Results}, there is a significant difference in the precision values with the STN-YOLO model as compared to the baseline model. The precision metrics reveal a substantial improvement, showcasing the model's enhanced ability to accurately detect objects.  Visually, we can see this difference in the Figure \ref{subfig:melonflower} where in the YOLO detections of the Melonflower dataset, the model detects the melon flower with an IOU of $0.50$ and the STN-YOLO detections of the Melonflower dataset has a better improved detection with an IOU value of $0.60$. This difference not only emphasizes the effectiveness of the STN-YOLO model but also underscores its potential for applications where precision is a critical performance indicator. Since the MelonFlower dataset had the most spatial variance among all datasets evaluated in this work, the STN-YOLO model exhibited notably improved performance, as evidenced by the precision values and other object detection metrics presented in Table \ref{tab:Benchmark Datasets Results}.

\section{Conclusion}
This work focused on integrating STN with YOLO, creating the STN-YOLO model to address spatial invariance challenges. The model aimed to enhance plant image detection quality and can be used for downstream applications such as phenotypic feature extraction. Results demonstrated that STN improved model robustness and reduced number of false positives in the datasets as indicated by the higher precision scores. The STN-YOLO model improved performance on benchmark datasets, showcasing potential in handling real-world spatial transformations. Future work includes integrating STNs into other object detection models (\textit{e.g.}, future versions of YOLO \cite{yolov10}), developing new objective functions to improve the learning of the model with STNs, expanding our PGP dataset to include more images across multiple crops with different image conditions (\textit{e.g.,} illumination), and incorporating the near infrared channel. 

\section{Acknowledgement}

This material is based upon work supported by Texas A\&M AgriLife. Portions of this research were conducted with the advanced
computing resources provided by Texas A\&M High Performance Research Computing.

\balance
\bibliographystyle{IEEEtran}
\bibliography{main.bib}

\end{document}